\documentclass{article}

    \PassOptionsToPackage{numbers, compress}{natbib}

\newcommand{\blist}{\begin{itemize}\topsep=0.0pt \parsep=0pt \itemsep=-0.1pt}
\newcommand{\elist}{\end{itemize}}

\newcommand{\nlist}{\begin{enumerate}\topsep=0.0pt \parsep=0pt \itemsep=-0.1pt}
\newcommand{\nelist}{\end{enumerate}}
    \usepackage[preprint]{neurips_2024}

\usepackage[utf8]{inputenc} 
\usepackage[T1]{fontenc}    
\usepackage{hyperref}       
\usepackage{url}            
\usepackage{booktabs}       
\usepackage{amsfonts}       
\usepackage{nicefrac}       
\usepackage{microtype}      
\usepackage{xcolor}         

\usepackage{algpseudocode}
\usepackage[ruled, linesnumbered]{algorithm2e}
\usepackage{amsmath}
\usepackage{graphicx}
\usepackage{etoolbox,lineno}

\title{Superfast Selection for Decision Tree Algorithms\thanks{Authors partially supported by NSF Grant IIS 1910131 and 1718945}}

\author{%
  Huaduo Wang \\
  The University of Texas at Dallas \\
  Richardson, TX 75080 \\
  \texttt{huaduo.wang@utdallas.edu} \\
  \And
  Gopal Gupta \\
  The University of Texas at Dallas \\
  Richardson, TX 75080 \\
  \texttt{gupta@utdallas.edu} \\
}

\begin{document}
\nolinenumbers

\maketitle

\begin{abstract}
  We present a novel and systematic method, called \textit{Superfast Selection}, for selecting the ``optimal split" for decision tree and feature selection algorithms over tabular data.  
  The method speeds up split selection on a single feature by lowering the time complexity, from O(M$\cdot$N) (using the standard selection methods) to O(M), where M represents the number of input examples and N the number of unique values. 
  Additionally, the need for pre-encoding, such as one-hot or integer encoding, for feature value heterogeneity is eliminated. 
  To demonstrate the efficiency of Superfast Selection, we empower the CART algorithm by integrating Superfast Selection into it, creating what we call \textit{Ultrafast Decision Tree} (UDT). 
  This enhancement enables UDT to complete the training process with a time complexity O(K$\cdot$M$^2$) (K is the number of features). 
  Additionally, the \textit{Training Only Once Tuning} enables UDT to avoid the repetitive training process required to find the optimal hyper-parameter.
  Experiments show that the UDT can finish a single training on KDD99-10\% dataset (494K examples with 41 features) within 1 second and tuning with 214.8 sets of hyper-parameters within 0.25 second on a laptop.
\end{abstract}


\section{Introduction}

The field of Machine Learning has been rapidly evolving with the development of complex models that can tackle sophisticated tasks \cite{sarker2021deep}. 
These advancements represent significant milestones, showing the field's capacity for innovation and its relentless pursuit of cutting-edge technology. 
However, despite the allure of these elaborate models,
the backbone of machine learning applications in real-world scenarios remains rooted in more lightweight and 
simple models \cite{rudin2019}. 
These models not only offer practical solutions to a wide array of problems but also underscore the importance of accessibility, interpretability, and computational efficiency in the vast landscape of machine learning \cite{rudin2021interpretable}. 

In machine learning, decision tree algorithms stand out for their simplicity and interpretability, making them a staple in both academic research and practical applications. 
As the volume and complexity of data continue to grow, optimizing these algorithms for enhanced efficiency, accuracy, and scalability becomes increasingly vital. 
This has led to the development of a variety of optimization techniques, aimed at refining every aspect of decision tree learning—from the construction phase to the model's final deployment. 
Techniques such as pruning \cite{quinlan1987}
feature selection \cite{quinlan1993}, and tree ensemble methods \cite{bishop2006}
have helped greatly improve the performance of decision trees.  
Advancements in splitting criteria \cite{kass1980}, parallel/distributed computing \cite{chen2016}, and incremental learning have allowed these algorithms to handle \cite{domingos2000} large-scale data analyses.

Split selection is a fundamental aspect of training decision tree algorithms. 
It involves choosing the optimal value and feature to divide the data at each node of the tree. 
This process is crucial for effectively building the tree's structure, which is aimed at solving classification or regression problems by learning simple decision rules inferred from the training data. 
Most of the research in this field is about heuristic metrics that guide selection choices \cite{blanc2021decision}. 
The generic approach to selecting splits involves comprehensively calculating the heuristic scores for all candidate splits and then choosing the one with the highest score.
However, the O(N$^2$)  complexity of the selection method constraints model training, where N is the number of training instances.  
%
%
The research field of systematically computing the heuristic scores still remains under-explored.

We propose a novel selection algorithm---\textit{Superfast Selection}---to compute the heuristic score of all possible candidate splits and make selection systematically. 
Superfast Selection algorithm speeds up the decision tree training process by lowering the time complexity of the split selection process. 
We enhance CART  (classification and regression tree) \cite{breiman1984} algorithm by integrating Superfast Selection and \textit{Training Only Once Tuning} to develop Ultrafast Decision Tree, demonstrating the potency of Superfast Selection.

\section{Superfast Selection Algorithm}

\noindent\textbf{Background:}
%
\textit{Split selection} in a decision tree is governed by specific criteria meant to maximize the homogeneity of the resultant subsets after the splits. 
Different algorithms may use different metrics for this purpose. The primary goal is to make decisions at each node that best separates the data into classes (for classification tasks) or minimize variance (for regression tasks). 
A split selection process of a single tree node typically works as follows:

\vspace{-0.1in}
\blist 
    \item 
For each candidate feature, find all potential splits. For categorical features, it might involve grouping categories together, while for numerical features, it involes finding thresholds.
    \item 
Calculate the chosen metric (like Gini Impurity \cite{breiman1984}, Information Gain \cite{quinlan1986}, variance, etc.) for each candidate split.
\item 
Select the split that results in the highest score of the chosen metric.
\elist 
\vspace{-0.1in}


Superfast Selection is an algorithm framework that is compatible with the most commonly used split criteria, like Gini Impurity, Information Gain, Chi-Square \cite{pearson1900}, and variance (for regression tasks), etc.
The most common data a decision tree deals with is tabular data, and the input features generally fall into two categories: numerical features and categorical features. 
Superfast Selection can deal with both numerical and categorical features as well as hybrid features (a feature that contains both numerical values and categorical values simultaneously) without the need for any pre-encoding such as one-hot encoding or integer encoding.

\smallskip\noindent\textbf{Comparison with Generic Split Selection:}
%
The generic split selection method outlined here serves as a broad abstraction of the widely used split selection process in decision tree algorithms. 
The generic split selection algorithm and Superfast Selection algorithm on a single feature are summarized as Algorithm \ref{algo:ogss} and Algorithm \ref{algo:osss}, respectively. 
In Algorithms \ref{algo:ogss} and \ref{algo:osss}, M is the number of examples, N is the number of unique values of the feature, and C is the number of label classes. 

\vspace{-0.1in}
\IncMargin{1.5em}
\begin{algorithm}[!h]
\small
\caption{Overview of Generic Split Selection}
\label{algo:ogss}
\SetKwInOut{Input}{input}
\SetKwInOut{Output}{output}
\SetKwFunction{bestonattr}{generic\_selection\_on\_single\_feat}
\DontPrintSemicolon
\SetKwProg{Fn}{Function}{}{end}
\Fn{\bestonattr}{
    scan feature values to get a unique feature value set $V$ \Comment{O(M)}  \;
    // loop N times to compute heuristics of all splits \Comment{N=$|V|$}\;
    \For {$x \in V$} { 
        scan all feature values and example labels \Comment{O(M)} \;
        compute heuristic scores of split of $x$ \;
    }
    return split with the highest heuristic
}
\end{algorithm}

\begin{algorithm}[!h]
\small
\caption{Overview of Superfast Selection}
\label{algo:osss}
\SetKwInOut{Input}{input}
\SetKwInOut{Output}{output}
\SetKwFunction{bestonattr}{superfast\_selection\_on\_single\_feat}
\DontPrintSemicolon
\SetKwProg{Fn}{Function}{}{end}
\Fn{\bestonattr}{
    scan feature values and labels with one pass \Comment{O(M)} \;
    prepare intermediate statistics of size O(N$\cdot$C) and unique feature value set $V$ \Comment{N=$|V|$} \;
    // loop N times to compute heuristics of all splits\;
    \For {$x \in V$} {
        compute heuristic scores of split of $x$ \Comment{O(C)}\;
    }
    return split with the highest heuristic
}
\end{algorithm}
\vspace{-0.1in}

Generic selection (Algorithm \ref{algo:ogss}) first collects the unique value set of the feature with O(M) time cost, then loops N times to compute the heuristic scores of all candidate splits. 
Inside each loop, the algorithm scans all values of the feature and example labels with O(M) time cost and then computes the heuristic score. 
Thus, the time complexity of generic split selection on a single feature is O(M$\cdot$N).

Superfast Selection (Algorithm \ref{algo:osss}) first scans all values of the feature and example labels and prepares intermediate statistics information of size O(N$\cdot$C) and the unique value set of the feature with time cost O(N$\cdot$C). 
Then, the algorithm loops N times to compute the heuristic scores of all splits of each unique value. 
Inside each loop, it only needs O(C) time cost to compute the heuristic of all splits of this value by using the intermediate statistics. 
Hence, the time complexity of Superfast Selection on a single feature is O(M+N$\cdot$C). 
Generally, the number of label classes C is considered a constant number. In this case, the complexity can be simplified as O(M). 
In contrast, the time complexity of the generic split selection is O(M$\cdot$N).

\noindent\textbf{Handling Missing Values:}
%
Missing data is so common that it is the norm rather than an exception \cite{kang2013}. There are many ways to deal with the missing values that have been suggested: 
(i) removing the rows/columns,
(ii) imputation that replaces missing values with mean, median, or mode of observed values in that column,
(iii) predicting the missing value with machine learning or data mining techniques using the training data.
However, these methods essentially change the original data by either adding extra information or discarding defective samples. To keep the algorithm simple and data authenticity, Superfast Selection deals with missing values by leaving them untouched
without losing or adding any information.

\noindent\textbf{Split Candidates:} 
%
%
The generic split selection procedure generates numerical comparison (``$\leq$'' and ``$>$'') splits for numerical features and generates equality (``$=$'' and ``$\ne$'') comparison splits for categorical features. 
Superfast Selection algorithm handles all the values of a categorical feature as categorical values, same as generic split selection. 
To deal with hybrid features, Superfast Selection reads each value of a numerical or hybrid feature as a number first, converting it to a categorical value if the conversion fails. 
Numerical comparison (``$\leq$'' and ``$>$'') split candidates would be generated for each numerical value, and equality comparison (``$=$'' and ``$\ne$'') split candidates would be generated for each categorical value. 
In this way, Superfast Selection algorithm enumerates all the values in the features and then exhaustively computes their heuristic scores.

\noindent\textbf{Comparison Assumption:}
%
Superfast Selection algorithm employs a carefully designed comparison operator that can directly handle hybrid features without use of pre-encoding, such as one-hot encoding or integer encoding. 
Equality of two same-type values (categorical or numerical) is straightforward as commonsense dictates: they are equal if they are identical, otherwise, they are unequal. 
The equality of different types of values is always false, therefore their inequality is always true. 
The numerical comparison between a categorical value and a numerical value is always false. Table 3 shows the evaluations of different comparisons between 10 and `cat'.

\begin{table}[]
\centering
\footnotesize
\begin{tabular}{cc}
    \begin{minipage}{.4\linewidth}
        \setlength{\tabcolsep}{5pt}
        \centering
        \begin{tabular}{|c|c|c|c|c|c|c|c|c|}
        \cline{1-9}
          & \multicolumn{8}{c|}{\textbf{feature values}}\\
        \cline{1-9}
        $\mathbf{E_a}$ & 3 & 4 & 4 & 5 & x & x & y & \\
        \cline{1-9}
        $\mathbf{E_b}$ & 1 & 1 & 2 & 2 & 3 & y & y & z  \\
        \cline{1-9}
        $\mathbf{E_c}$ & 3 & 4 & 4 & 5 & 5 & z & z &  \\
        \cline{1-9}
        \end{tabular}
        \label{tbl:example1}
        \smallskip 
        \caption{Examples and feature values}
        
        \centering
        \begin{tabular}{|c|c|c|c|c|c|c|c|c|}
        \cline{1-9}
          & \multicolumn{8}{c|}{\textbf{feature value statistics}}\\
        \cline{1-9}
        \textbf{value} & 1 & 2 & 3 & 4 & 5 & x & y & z \\
        \cline{1-9}
        $\mathbf{cnt_a}$ & 0 & 0 & 1 & 2 & 1 & 2 & 1 & 0 \\
        \cline{1-9}
        $\mathbf{pfs_a}$ & 0 & 0 & 1 & 3 & 4 & na & na & na \\
        \cline{1-9}
        $\mathbf{cnt_b}$ & 2 & 2 & 1 & 0 & 0 & 0 & 2 & 1 \\
        \cline{1-9}
        $\mathbf{pfs_b}$ & 2 & 4 & 5 & 5 & 5 & na & na & na \\
        \cline{1-9}
        $\mathbf{cnt_c}$ & 0 & 0 & 1 & 2 & 2 & 0 & 0 & 2 \\
        \cline{1-9}
        $\mathbf{pfs_b}$ & 0 & 0 & 1 & 3 & 5 & na & na & na \\
        \cline{1-9}
        \end{tabular}
        \label{tbl:example2}
        \caption{positive/negative count and prefix sum on each value}
    \end{minipage} &

    \begin{minipage}{.5\linewidth}
        \setlength{\tabcolsep}{2pt}
        \centering
        \begin{tabular}{|c|c|c|c|}
        \cline{1-4}
        \textbf{comparison}  & \textbf{evaluation} & \textbf{comparison}  & \textbf{evaluation} \\
        \cline{1-4}
        10 $=$ `cat' & False & 10 $\neq$ `cat' & True\\ 
        \cline{1-4}
        10 $\le$ `cat' & False & 10 $>$ `cat' & False \\ 
        \cline{1-4}
        \end{tabular}
        \label{tbl:example3}
        \medskip 
        \caption{Comparing numerical \& categorical values}
        
        \centering
        \begin{tabular}{|c|c|c|c|c|c|c|c|c|}
        \cline{1-9}
         & \multicolumn{8}{c|}{\textbf{heuristic}}  \\
        \cline{1-9}
        \textbf{val} & 1 & 2 & 3 & 4 & 5 & x & y & z \\
        \cline{1-9}
        \textbf{$\leq$ val} & -0.99 & \textbf{-0.87} & -0.97 & -1.08 & -1.06 & na & na & na \\
        \cline{1-9}
        \textbf{$>$ val} & -1.06 & -0.96 & -0.92 & -1.04 & -1.15 & na & na & na \\
        \cline{1-9}
        \textbf{$=$ val} & na & na & na & na & na & -0.98 & -1.03 & -1.01  \\
        \cline{1-9}
        \end{tabular}
        \label{tbl:example4}
        \smallskip 
        \caption{The heuristic on feature values of given example}
    \end{minipage} 
\end{tabular}
\vspace{-0.2in}
\end{table}

\noindent\textbf{Compute Heuristic with Prefix-Sum:}
%
Superfast Selection algorithm utilizes the \textit{prefix-sum} technique to speed up the heuristic computing process of a feature. 
The carefully designed comparison operator is a prerequisite for using the prefix-sum technique. 
Because it makes all evaluation results of split candidates fall into only two sets (positive and negative) in a consolidated way.

Here we use Information Gain \cite{quinlan1986} as an example heuristic to show how Superfast Selection works, while other heuristics, like Gini Index \cite{quinlan1986}, Gini Impurity \cite{breiman1984}, and Chi-Square \cite{pearson1900}, can also be integrated into our Superfast Selection method:

\vspace{-0.15in}
\begin{equation}
\footnotesize
    IG(T,a)=H(T)-H(T|a)
\label{eq:ig}
\end{equation}

\vspace{-0.12in}

\noindent where $H(T)$ is the entropy of the training set $T$ while $H(T|a)$ is the cross-entropy of the training set on feature $a$. 
In practice, $H(T)$ can be ignored for comparison because all the candidates would have the same entropy value. 
Therefore, only $H(T|a)$ is needed for comparison. 
Then the information gain computation for comparison is simplified as:

\vspace{-0.2in} 
\begin{equation}
\footnotesize
    {\frac{1}{M} \sum\limits_{i=1}^{C}(p_i\cdot log\frac{p_i}{\Sigma_{j=1}^{C}p_j})_{p_i>0}+ \frac{1}{M} \sum\limits_{i=1}^{C}(n_i\cdot log\frac{n_i}{\Sigma_{j=1}^{C}n_j})_{n_i>0}}
    \label{eq:sig}
\end{equation}
\vspace{-0.2in}

\noindent where $p_i,n_i$ denotes the number of positive predictions and the number of negative predictions for the examples of $class_i$ with binary splitting, respectively; 
C denotes the number of label classes, and M denotes the number of examples. 
In Superfast Selection, this equation costs  O(C) time to compute.

\vspace{-0.1in}
\begin{algorithm}[!h]
\small
\caption{Simplified information gain function}
\label{algo:sigf}
\SetKwInOut{Input}{input}
\SetKwInOut{Output}{output}

\SetKwFunction{bestonattr}{best\_split\_on\_attr}
\SetKwFunction{cntsort}{counting\_sort}
\SetKwFunction{h}{heuristic}
\SetKwFunction{log}{log}
\SetKwFunction{ssum}{sum}

\DontPrintSemicolon

\Input{\;$\textit{pos}$: the number of positive predictions of each class with binary split\; $\textit{neg}$: the number of negative predictions of each class with binary split}
\Output{$\textit{ret}$: heuristic score}
\SetKwProg{Fn}{Function}{}{end}

\Fn{\h{$\textit{pos},\textit{neg}$}}{
    $\textit{tot}_p,\ \textit{tot}_n \gets$ \ssum{$\textit{pos}$}, \ssum{$\textit{neg}$}\;
    $\textit{ret},\ \textit{tot} \gets 0,\ \textit{tot}_p+\textit{tot}_n$\;
    \For{$\textit{p} \in \textit{pos}$}{
        \If{$\textit{p}>0$}{
            $\textit{ret} \gets \textit{ret} + \frac{\textit{p}}{\textit{tot}} \cdot$ \log{$\frac{\textit{p}}{\textit{tot}_p}$}
        }
    }
    \For{$\textit{n} \in \textit{neg}$}{
        \If{$\textit{n}>0$}{
            $\textit{ret} \gets \textit{ret} + \frac{\textit{n}}{\textit{tot}} \cdot \log{$\frac{\textit{n}}{\textit{tot}_n}$}$
        }
    }
    \textbf{return} $ret$
}
\end{algorithm}
\vspace{-0.12in}

Equation \ref{eq:sig} can be realized as the heuristic function in Algorithm \ref{algo:sigf}.
Based on the above simplified heuristic function, the Superfast Selection can be summarized as the \texttt{best\_split\_on\_feat} function in Algorithm \ref{algo:sss}. 
In lines 2--9, the function initializes a statistics-collecting process for heuristic calculation. 
After gathering all statistics, the function calculates the prefix sum in lines 10--14. 
In lines 15--36, the function systematically computes the heuristic score for all the possible split candidates. 
Finally, the result is the best score with its corresponding split.

\begin{algorithm}[!h]
\small
\caption{Best Split function of Superfast Selection}
\label{algo:sss}
\SetKwInOut{Input}{input}
\SetKwInOut{Output}{output}

\SetKwFunction{bestonattr}{best\_split\_on\_feat}
\SetKwFunction{bestonallattr}{best\_split\_on\_all\_feats}
\SetKwFunction{cntsort}{counting\_sort}
\SetKwFunction{h}{heuristic}
\SetKwFunction{bestpair}{best\_pair}

\DontPrintSemicolon

\Input{$\textit{E}$: examples, $\textit{A}$: all features, $\textit{X}^\textit{A}$: pre-sorted numerical feature values of all features $\textit{A}$}
\Output{$\textit{h}$: the best heuristic score, $\textit{split}$: the split predicate with $h$}
\SetKwProg{Fn}{Function}{}{end}

\Fn{\bestonattr{{$\textit{E}$}, {$\textit{a}$}, {$\textit{X}$}}}{
    // initialize statistics collection for the input $\textit{E}$ on feature $\textit{a}$\;
    $\textit{C} \gets $ set of categorical values\;
    $\textit{Y} \gets $ set of label\;
    $\textit{cnt}_{n} \gets $ count of label $\textit{y}$, numerical feature value $\textit{x}$\;
    $\textit{cnt}_{c} \gets $ count of label $\textit{y}$, categorical feature value $\textit{x}$\;
    $\textit{tot}_{n} \gets $ total count of numerical feature values of label $\textit{y}$\;
    $\textit{tot}_{c} \gets $ total count of categorical feature values of label $\textit{y}$\;
    // end of statistics collection\;
    
    \For{$\textit{y} \in \textit{Y}$}{
        \For{$j\leftarrow 1$ \KwTo $|\textit{X}|$}{
            $\textit{cnt}_n[\textit{y},\textit{X}_{j}]\gets \textit{cnt}_n[\textit{y},\textit{X}_{j}]+\textit{cnt}_n[\textit{y},\textit{X}_{j-1}]$\;
        }
    }
    
    \For{$\textit{x} \in \textit{X}$}{
        $\textit{pos}, \textit{neg} \gets list(), list()$\;
        \For{$\textit{y} \in \textit{Y}$}{
            $\textit{pos}.\textit{add}(\textit{cnt}_n[\textit{y},\textit{x}])$\;
            $\textit{neg}.\textit{add}(\textit{tot}_n[\textit{y}]-\textit{cnt}_n[\textit{y},\textit{x}]+\textit{tot}_c[\textit{y}])$\;
        }
        $\textit{score}[(\textit{a},\leq,\textit{x})] \gets$ \h{$\textit{pos}, \textit{neg}$}\;
        $\textit{pos}, \textit{neg} \gets list(), list()$\;
        \For{$\textit{y} \in \textit{Y}$}{
            $\textit{pos}.\textit{add}(\textit{tot}_n[\textit{y}]-\textit{cnt}_n[\textit{y},\textit{x}])$\;
            $\textit{neg}.\textit{add}(\textit{cnt}_n[\textit{y},\textit{x}]+\textit{tot}_c[\textit{y}])$\;
        }
        $\textit{score}[(\textit{a},>,\textit{x})] \gets$ \h{$\textit{pos}, \textit{neg}$}\;
    }
    \For{$\textit{x} \in \textit{C}$}{
        $\textit{pos}, \textit{neg} \gets list(), list()$\;
        \For{$\textit{y} \in \textit{Y}$}{
                $\textit{pos}.\textit{add}(\textit{cnt}_c[\textit{y},\textit{x}])$\;
                $\textit{neg}.\textit{add}(\textit{tot}_c[\textit{y}]-\textit{cnt}_c[\textit{y},\textit{x}]+\textit{tot}_n[\textit{y}])$\;
            }
        $\textit{score}[(\textit{a},=,\textit{x})] \gets$ \h{$\textit{pos}, \textit{neg}$}\;
    }
    $\textit{h}, \textit{split} \gets$ \bestpair{$\textit{score}$}\;
    \textbf{return} $\textit{h}, \textit{split}$
}
\Fn{\bestonallattr{{$\textit{E}$}, {$\textit{A}$}, {$\textit{X}^\textit{A}$}}}{
    \For{$\textit{a} \in \textit{A}$}{
        $\textit{h}, \textit{split} \gets$ \bestonattr{{$\textit{E}$}, {$\textit{a}$}, {$\textit{X}_a^\textit{A}$}}\;
        $\textit{sccore}[\textit{split}] \gets \textit{h}$\;
    }
    $\textit{best\_h}, \textit{best\_split} \gets$ \bestpair{$\textit{score}$}\;
    \textbf{return} $\textit{best\_split}$\;
}
\end{algorithm}

Here is an example of the heuristic calculation process in Table 1: there are 7 examples with the label ``a'', 8 examples with the label ``b'', and 7 examples with the label ``c''. 
Their feature values are listed in the top table. 
Their corresponding count $\textup{cnt}_x$ and prefix sum $\textup{pfx}_x$ in Table 2 have been computed as lines 2--9 in Algorithm \ref{algo:sss}. 
Then, the heuristic scores of all possible splits in Table 4 have been calculated as lines 15--36. 
Finally, -0.87 with val $\leq$ 2 is returned as the result of the \texttt{best\_split\_on\_feat} function.

\noindent\textbf{Complexity:}
%
As mentioned above in this section, the time complexity of Superfast Selection on a single feature is O(M+N$\cdot$C) and can be simplified in practice to O(M). 
However, the \texttt{best\_split\_on\_feat} function in Algorithm \ref{algo:sss} takes a pre-sorted number list as an input parameter. 
Typically, the sorting process of a set of numbers with size O(M) can be finished within O(MlogM) time. 
However, the sorting results and cost can be shared across the whole tree-building process since the numbers only need to be sorted once at the beginning. 
The overhead of preparing the sorted numbers for each split selection process is to maintain a list of sorted present numbers and filter out the split-out numbers during the node-splitting process which costs  O(M) time in total. 

\section{Ultrafast Decision Tree}

\textit{Classification and Regression Tree} (CART) is a good example to illustrate the power of Superfast Selection. 
We assume readers are already familiar with the CART algorithm since it is well-known and widely used. Our
\textit{Ultrafast Decision Tree} (UDT) is summarized in Algorithm \ref{algo:utd}. 
To optimally apply the Superfast Selection algorithm on the UDT, all numerical values of each feature are sorted at the initial stage of the tree-building process in line 2. 
In the node-splitting process, to maintain the sorted feature values, a hash table does help to identify if a value appears in a child node for each feature. 
In lines 15--16, the function \texttt{filter\_sorted\_nums} filters out the numerical values that are not included in $node.X^A$ for each feature. 
The sorted feature values $node.X^A$ have been divided into $X^A_+$ and $X^A_-$ based on the evaluation of examples.
In this way, the cost of sorting numerical values is distributed across the whole tree-building process. 

\begin{algorithm}[!h]
\small
\caption{Ultrafast Decision Tree algorithm}
\label{algo:utd}
\SetKwInOut{Input}{input}
\SetKwInOut{Output}{output}
\Input{$\textit{E}$: examples, $\textit{A}$: all candidate features}
\Output{$\textit{root}$: the root of the decision tree}
\SetKwFunction{buildtree}{build\_tree}
\SetKwFunction{bestsplit}{best\_split\_on\_all\_attrs}
\SetKwFunction{evaluate}{eval\_and\_split}
\SetKwFunction{bestonallattr}{best\_split\_on\_all\_feats}
\SetKwFunction{splitnode}{split\_node}
\SetKwFunction{generate}{generate\_label}
\SetKwFunction{treenode}{node}
\SetKwFunction{init}{init}
\SetKwFunction{pop}{pop}
\SetKwFunction{push}{push}
\SetKwFunction{filter}{filter\_sorted\_nums}

\DontPrintSemicolon

\SetKwProg{Fn}{Function}{}{end}

\Fn{\buildtree{{$\textit{E}$}, {$\textit{A}$}}}{
    ${\textit X^A} \gets$ sorted numerical values of all features of ${\textit E}$\;
    $\textit{root} \gets$ \treenode{{$\textit{E}$}, {${\textit X^A}$}}\; 
    $\textit{Q}$.\init{$\textit{root}$}\;
    \While {\textup{$\textit{Q}$} \textbf{is not} empty} {
        $\textit{node} \gets$ $\textit{Q}$.\pop{}\;
        \splitnode{$\textit{node}$}\;
    }
    \textbf{return} $\textit{root}$\;
}

\Fn{\splitnode{$\textit{node}$}}{
    $\textit{node.split} \gets $ \bestonallattr{{$\textit{node.E}$}, {$\textit{node.X$^\textit{A}$}$}}\;
    $\textit{node.y} \gets $ \generate{{$\textit{node.E}$}}\;
    $\textit{E}_\textit{+}, \textit{E}_\textit{-} \gets $\evaluate{{$\textit{node.E}$}, {$\textit{node.split}$}}\;
    $\textit{X}^\textit{A}_\textit{+} \gets$ \filter{{$\textit{node.X$^\textit{A}$}$}, {$\textit{E}_\textit{+}$}}\;
    $\textit{X}^\textit{A}_\textit{-} \gets$ \filter{{$\textit{node.X$^\textit{A}$}$}, {$\textit{E}_\textit{-}$}}\;
    $\textit{node.child$_+$} \gets$ \treenode{{$\textit{E}_\textit{+}$}, {$\textit{X}^\textit{A}_\textit{+}$}}\;
    $\textit{node.child$_-$} \gets$ \treenode{{$\textit{E}_\textit{-}$}, {$\textit{X}^\textit{A}_\textit{-}$}}\;
    $\textit{Q}$.\push{$\textit{node.child$_+$}$}\;
    $\textit{Q}$.\push{$\textit{node.child$_-$}$}\;
}

\end{algorithm}

\smallskip\noindent\textbf{Label Split for Regression Tasks:}
%
CART uses the sum of square error (SSE) as the criterion of label splitting with numerical values:

\vspace{-0.15in}
\begin{equation}
\begin{split}
\scriptsize
    SSE=\textstyle\sum\limits_{i\in S_1}y_i^2 + \textstyle\sum\limits_{i\in S_2}y_i^2 - \frac{1}{|S_1|}(\textstyle\sum\limits_{i\in S_1}y_i)^2 - \frac{1}{|S_2|}(\textstyle\sum\limits_{i\in S_2}y_i)^2
    \label{eq:sse}
\end{split}
\end{equation}

In the above formula, $S_1$ and $S_2$ are the two label sets the split creates. 
$\textstyle\sum\limits_{i\in S_1}y_i^2 + \textstyle\sum\limits_{i\in S_2}y_i^2$ can be ignored for the same label set due to having the same value. 
Then, numerical label selection can be optimized with prefix-sum as in Algorithm \ref{algo:labelselection}. 
If the number of examples is M, the complexity of finding the best split for the label is O(M). 
Note the number of classes in the split selection process is always two, the overhead of splitting the label won't add extra cost to the time complexity of the tree-building process. 

\begin{algorithm}[!h]

\small
\caption{Numerical Label Selection Function}
\label{algo:labelselection}
\SetKwInOut{Input}{input}
\SetKwInOut{Output}{output}

\SetKwFunction{bestonattr}{best\_split\_on\_attr}
\SetKwFunction{cntsort}{counting\_sort}
\SetKwFunction{h}{sse}
\SetKwFunction{bestsplit}{best\_label\_split}
\SetKwFunction{bestpair}{best\_pair}
\SetKwFunction{filter}{filter\_sorted\_nums}

\DontPrintSemicolon

\Input{$\textit{node}$: tree node, $\textit{L}$: pre-sorted numerical labels}
\Output{$\textit{y}$: best numerical split on label}
\SetKwProg{Fn}{Function}{}{end}
\Fn{\bestsplit{$\textit{node}$, $\textit{L}$}}{
    $\textit{Y} \gets $ \filter{{$\textit{L}$}, {$\textit{node.E}$}} \;
    $\textit{tot} \gets $ sum of all labels of $\textit{node.E}$\;
    $\textit{cnt} \gets $ count of each label $\textit{y} \in \textit {Y}$\;
    \For{$\textit{i}\leftarrow 1$ \KwTo $|\textit{Y}|$}{
        $\textit{cnt}[\textit{Y$_\textit{i}$}] \gets \textit{cnt}[\textit{Y$_\textit{i}$}] + \textit{cnt}[\textit{Y$_\textit{i-1}$}]$\;
    }
    $\textit{Y$_\textit{-1}$},\ \textit{cnt}[\textit{Y$_\textit{-1}$}],\ \textit{sum}_{y} \gets 0,\ 0,\ 0$\;
    
    \For{$\textit{i} \leftarrow 0$ \KwTo $|\textit{Y}|-1$}{
        $\textit{sum}_\textit{y} \gets \textit{sum}_\textit{y} + \textit{Y}_\textit{i} \cdot (\textit{cnt}[\textit{Y$_\textit{i}$}] - \textit{cnt}[\textit{Y$_\textit{i-1}$}])$\;
        $\textit{score}[\textit{Y}_\textit{i}] \gets -\textit{sum}_\textit{y}^2 \cdot \textit{cnt}[\textit{Y$_\textit{i}$}]^{-1} - (\textit{tot}-\textit{sum}_\textit{y})^2 \cdot (|\textit{node.E}|-\textit{cnt}[\textit{Y$_\textit{i}$}])^{-1}$\;
    }
    $\textit{sse},\ \textit{y} \gets \bestpair{\textit{score}}$\;
    \textbf{return} $\textit{y}$\;
}
\end{algorithm}

\noindent\textbf{Training Only Once Tuning:}
Decision tree models, while highly intuitive and interpretable, are particularly prone to overfitting on the training set. 
This susceptibility originates from the inherent flexibility and ability to create complex decision boundaries that perfectly classify the training data. 
Many hyper-parameters are used for tuning decision tree models to mitigate overfitting, the most commonly used are \textit{Maximum Depth}, \textit{Minimum Samples Split}, and \textit{Maximum Features}, etc. 
Ultrafast Decision Tree algorithm currently employs \textit{Maximum Depth} and \textit{Minimum Samples Split} as the criteria for tuning the model. 
With these two hyper-parameters, it's not necessary to train multiple times for tuning models with different parameters, since the tree would be built with exactly the same pattern if training data and candidate features were not changed. 
The training process of UDT also determines the label of each node in the node-splitting sub-process. 
After training a full decision tree, the hyper-parameters would be used in the \textit{predict} function in Algorithm \ref{algo:prediction} to return results on non-leaf nodes during the tuning process. 
Finally, the tree model will be pruned based on the optimal evaluation result among all hyper-parameter settings.

\begin{algorithm}[!h]
\small
\caption{Ultrafast Decision Tree Predict Function}
\label{algo:prediction}
\SetKwInOut{Input}{input}
\SetKwInOut{Output}{output}

\SetKwFunction{bestonattr}{best\_split\_on\_attr}
\SetKwFunction{cntsort}{counting\_sort}
\SetKwFunction{h}{sse}
\SetKwFunction{bestsplit}{best\_label\_split}
\SetKwFunction{eval}{eval}
\SetKwFunction{filter}{filter\_sorted\_nums}
\SetKwFunction{predict}{predict}

\DontPrintSemicolon

\Input{$\textit{node}$: tree node, $\textit{L}$: pre-sorted numerical labels}
\Output{$\textit{node.y}$: prediction label}
\SetKwProg{Fn}{Function}{}{end}
\Fn{\predict{{$\textit{root}$}, {$\textit{x}$}, {$\textit{depth$_\textit{max}$}$}, {$\textit{split$_\textit{min}$}$}}}{
    $\textit{node} \gets \textit{root}$\;
    \For{$\textit{i} \gets 1$ \KwTo $\textit{depth$_\textit{max}$} - 1$}{
        \If{$\textit{node}$ \textbf{is} leaf \textbf{or} |$\textit{node.E}$| $< \textit{split$_\textit{min}$}$}{
            \textbf{break}
        }
        \uIf{\eval{{$\textit{node.split}$}, {$\textit{x}$}}}{
            $\textit{node} \gets \textit{node.child}_\textit{+}$
        }
        \Else {
            $\textit{node} \gets \textit{node.child}_\textit{-}$
        }
    }
    \textbf{return} $\textit{node.y}$\;
}
\end{algorithm}

\noindent\textbf{Complexity:}
%
Assuming the average case results in a balanced tree with a depth of O(logM).
The sorting of numerical values for K candidate features can be completed in O(KMlogM) time where M is the number of examples, and the sorting result can be reused throughout the entire tree-building process.
Therefore, the average bound of time complexity for building a UDT is $\Theta$(KMlogM), given that there are K candidate features and the cost for selecting a split on a single feature is O(M). 
In the worst-case scenario for constructing a binary decision tree, the outcome is an unbalanced tree with O(M) depth. 
The Ultrafast Decision Tree algorithm powered by Superfast Selection can build decision trees within O(KM$^2$) time in total. 

\noindent\textbf{Limitations:} Our Superfast Selection method can be used in a wide variety of situations where decision trees are used. Our method significantly speeds up the training process. 
However, like the generic selection method, our method remains insensitive to linear relationships between features and labels as the split selection primarily focuses on local decisions within each node.

\section{Experiment}

We report on experiments that we conducted with Superfast Selection algorithm and Ultrafast Decision Tree (UDT) algorithm. 
First, we compare Superfast Selection with the generic selection method on a single feature. 
In the comparison experiment, we performed two selection algorithms on the “credit\_card\_fraud” dataset which contains 1 million samples with 7 features.  
Both algorithms are implemented in C++ and the testing machine is a Macbook Air M2 silicon with 8GB memory. 
We performed each selection algorithm 10 times on a single feature with different sizes that range from 10K to 100K then reported the average time cost for each size in Table 5. 

\begin{table}[]
\centering
\footnotesize
\begin{tabular}{cc}
    \begin{minipage}{.4\linewidth}
        \setlength{\tabcolsep}{5pt}
        \centering
        \begin{tabular}{|c|c|c|}
        \cline{1-3}
        \textbf{data size} & \textbf{generic} & \textbf{superfast} \\
        \cline{1-3}
        10K & 1.8K & 4 \\
        \cline{1-3}
        20K & 6.8K & 10 \\
        \cline{1-3}
        30K & 15K & 15 \\
        \cline{1-3}
        40K & 27K & 23 \\
        \cline{1-3}
        50K & 42K & 28 \\
        \cline{1-3}
        60K & 61K & 32 \\
        \cline{1-3}
        70K & 83K & 38 \\
        \cline{1-3}
        80K & 110K & 44 \\
        \cline{1-3}
        90K & 142K & 51 \\
        \cline{1-3}
        100K & 178K & 58 \\
        \cline{1-3}
        \end{tabular}
        \label{tbl:example6}
    \end{minipage} &

    \begin{minipage}{.5\linewidth}
        \centering
        \includegraphics[width=5cm]{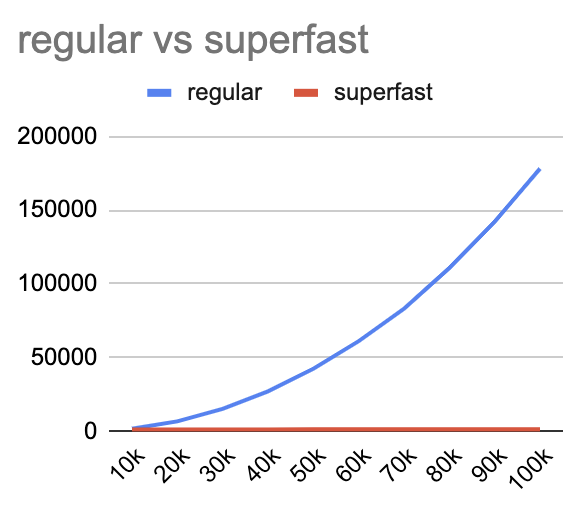}
        \label{fig1}
    \end{minipage}  
    
\end{tabular}
\caption{Time consumption (ms) comparison of generic and Superfast Selection on single feature}
\vspace{-0.15in}
\end{table}

To the best of our knowledge, no other decision tree algorithms have matched this level of efficiency.
Therefore, a comparative experiment with other DTs was considered unnecessary.
Next, we present the performance results of Ultrafast Decision Tree on 18 classification datasets of varying sizes. These datasets are taken from the UCI data repository and Kaggle and their details can be found elsewhere \cite{misc_adult_2, misc_default_of_credit_card_clients_350, misc_parkinsons_174, misc_online_shoppers_purchasing_intention_dataset_468, misc_statlog_(shuttle)_148, misc_wall-following_robot_navigation_data_194, misc_nursery_76, misc_page_blocks_classification_78, misc_weight_lifting_exercises_monitored_with_inertial_measurement_units_273, misc_letter_recognition_59, misc_optical_recognition_of_handwritten_digits_80, HeartDiseaseIndicators, ChurnModelling, misc_covertype_31, misc_kdd_cup_1999_data_130, credit_card_fraud}).
We treat 80\% of the data as the training set, 10\% as the validation set, and 10\% as the testing set. 
For each run, a full-fledged decision tree is trained without any limitation. 
Then, Training Only Once Tuning first evaluates on \textit{max\_depth} from 1 to the depth of the full decision tree, then evaluates on \textit{min\_split} from 0 to 4\% of the training set with a stepsize of 0.02\% (200 times for all datasets). 
The accuracy is calculated after pruning the full decision tree. 
Finally, a new decision tree is built based on the optimal hyper-parameter setting we tuned and a separate training time is reported.
We performed 10 cross-validation tests on these datasets and reported the average values of training time, tuning time, accuracies, numbers of nodes, and depths in Table \ref{tbl:example7}. 

Take ``churn modeling''(10K examples, 10 features, and 2 labels) as an example, we have 8K examples in the training set, 1K in the validation set, and 1K in the testing set. 
The training process of UDT takes 155 ms to create a full-fledged decision tree with 2165.6 nodes with 27.5 depth. Then, Training Only Once Tuning process takes 10ms to evaluate 227.5 sets of hyper-parameters (27.5 times for \textit{max\_depth} and 200 times for \textit{min\_split}) with 1K examples and pruning based on evaluation results. 
After pruning, we have a tuned tree with 224.4 nodes and 13.8 depth while achieving 0.85 accuracy on the test set. 
Finally, a separate training process with the tuned hyper-parameters costs 58 ms to build the same tuned tree.
In contrast, the generic tuning process repeats the training process 227.5 times and costs \textbf{16.8s} (\textit{Training Only Once Tuning} takes \textbf{10ms} for both tuning and pruning).

Superfast Selection algorithm eliminates the need for pre-encoding for the heterogeneity of feature values. The memory and time consumption for pre-encoding is saved. For example, one-hot encoding for ``credit card'' dataset needs about 39GB of memory and can not be performed on our 8GB testing machine. In contrast, training and tuning of UDT on ``credit card'' dataset consumes about 90MB of memory at peak.

\begin{table}[]
\centering
\footnotesize
{
\setlength{\tabcolsep}{1pt}
\begin{tabular}{|c|c|c|c|c|c|c|c|c|c|c|c|}
\cline{1-12}
\multicolumn{4}{|c|}{\textbf{Dataset}} & \multicolumn{4}{c|}{\textbf{Full tree}} & \multicolumn{4}{c|}{\textbf{Tuned tree}}  \\
 \cline{1-12}
\textbf{name} & \textbf{\#examples} & \textbf{\#feat.} & \textbf{\#label} & \textbf{node} & \textbf{depth} & \textbf{train{\scriptsize (ms)}} & \textbf{tune{\scriptsize (ms)}} & \textbf{acc.} & \textbf{node} & \textbf{depth}  & \textbf{train{\scriptsize (ms)}} \\ 
\cline{1-12}
adult & 32561 & 14 & 2 & 7238.6 & 61.4 & 586 & 50 & 0.86 & 427.6 & 27.1 & 180 \\
\cline{1-12}
credit card & 30000 & 23 & 2 & 6933 & 50.4 & 1340 & 52 & 0.82 & 138.4 & 13.5 & 299 \\
\cline{1-12}
rain in australia & 145460 & 23 & 3 & 26180.4 & 56.2 & 4229 & 288 & 0.83 & 701 & 17.3 & 627 \\
\cline{1-12}
parkinson & 765 & 753 & 2 & 66.8 & 10.5 & 611 & 2 & 0.80 & 30.2 & 6.0 & 363 \\
\cline{1-12}
intention & 12330 & 17 & 2 & 1496.8 & 31.6 & 170 & 6 & 0.90 & 105.6 & 11.5 & 50 \\
\cline{1-12}
shuttle & 58000 & 9 & 7 & 54.8 & 7.3 & 36 & 21 & 1.0 & 49.4 & 7.1 & 36 \\
\cline{1-12}
wall robot & 5456 & 24 & 4 & 51.2 & 10.6 & 70 & 2 & 0.99 & 40.2 & 9.3 & 66 \\
\cline{1-12}
nursery & 12960 & 8 & 5 & 436.6 & 16.0 & 18 & 5 & 1.0 & 425 & 15.9 & 18 \\
\cline{1-12}
page blocks & 5473 & 10 & 5 & 250.6 & 16.4 & 40 & 2 & 0.96 & 119 & 11.5 & 30 \\
\cline{1-12}
weight lifting & 4024 & 154 & 5 & 33.2 & 7.7 & 75 & 1 & 1.0 & 31.6 & 7.3 & 74 \\
\cline{1-12}
letter & 20000 & 16 & 26 & 3663 & 21.3 & 276 & 20 & 0.87 & 2563.6 & 19.4 & 272 \\
\cline{1-12}
nearest earth objects & 90836 & 7 & 2 & 11538 & 59.7 & 943 & 73 & 0.91 & 183.6 & 23.5 & 355 \\
\cline{1-12}
optidigits & 3823 & 64 & 10 & 398.4 & 12.6 & 121 & 2 & 0.89 & 309.6 & 11.4 & 99 \\
\cline{1-12}
heart disease indicators & 253680 & 21 & 2 & 54644 & 46.5 & 5802 & 453 & 0.91 & 658.8 & 18.0 & 717 \\
\cline{1-12}
credit card fraud & 1000000 & 7 & 2 & 66.2 & 8.0 & 5832 & 285 & 1.0 & 64.2 & 8.0 & 5836 \\
\cline{1-12}
churn modeling & 10000 & 10 & 2 & 2165.6 & 27.5 & 155 & 10 & 0.85 & 224.4 & 13.8 & 58 \\
\cline{1-12}
covertype & 581012 & 54 & 7 & 42756.4 & 41.3 & 16573 & 1023 & 0.94 & 42741 & 38.6 & 16740 \\
\cline{1-12}
kdd99-10\% & 494020 & 41 & 23 & 312.6 & 14.8 & 977 & 245 & 1.0 & 286.6 & 12.4 & 957 \\
\cline{1-12}
kdd99-full & 4898431 & 41 & 23 & 588.8 & 20.8 & 24926 & 3140 & 1.0 & 541.6 & 16.5 & 29625 \\
\cline{1-12}
\end{tabular}}
\medskip 
\caption{Ultrafast Decision Tree on various classification datasets}
\label{tbl:example7}
\end{table}

We also performed experiments on various regression datasets \cite{misc_bike_sharing_275, misc_wine_quality_186, misc_large-scale_wave_energy_farm_882, misc_appliances_energy_prediction_374, misc_electrical_grid_stability_simulated_data__471} and reported similar results in Table \ref{tbl:example8}.  Instead of accuracy, \textit{Root Mean Square Error} is used in the tuning process. \textit{Mean Absolute Error} and \textit{Root Mean Square Error} are reported for regression tasks.

\begin{table}[]
\centering
\footnotesize
{
\setlength{\tabcolsep}{1pt}
\begin{tabular}{|c|c|c|c|c|c|c|c|c|c|c|c|c|}
\cline{1-12}
\multicolumn{3}{|c|}{\textbf{Dataset}} & \multicolumn{4}{c|}{\textbf{Full tree}} & \multicolumn{5}{c|}{\textbf{Tuned tree}}  \\
 \cline{1-12}
\textbf{name} & \textbf{\#example} & \textbf{\#feat.} & \textbf{node} & \textbf{depth} & \textbf{train{\scriptsize (ms)}} & \textbf{tune{\scriptsize (ms)}} & \textbf{{\scriptsize MAE}} & \textbf{{\scriptsize RMSE}} & \textbf{node} & \textbf{depth}  & \textbf{train{\scriptsize (ms)}} \\ 
\cline{1-12}
bike\_sharing\_hour & 17379 & 12 & 26433.8 & 44.6 & 1216 & 26 & 39.1 & 64.2 & 3540.8 & 37.7 & 250 \\
\cline{1-12}
california\_housing & 20640 & 9 & 31704.8 & 33.4 & 1439 & 40 & 37773 & 57633.3 & 2110 & 24.2 & 349 \\
\cline{1-12}
wine\_quality & 6497 & 11 & 2792.2 & 27.3 & 180 & 6 & 0.48 & 0.83 & 2739 & 24.9 & 169 \\
\cline{1-12}
wave\_energy\_farm & 36043 & 148 & 17887.6 & 34.4 & 18630 & 147 & 2842.4 & 7979.9 & 16401.4 & 24.9 & 17035 \\
\cline{1-12}
applicances\_energy & 19735 & 27 & 17787.4 & 45.2 & 2576 & 40 & 41.5 & 94.6 & 11123.8 & 24.8 & 1803 \\
\cline{1-12}
\end{tabular}}
\medskip 
\caption{Ultrafast Decision Tree on various regression datasets}
\label{tbl:example8}
\end{table}

\section{Conclusions and Implications}


In this paper, we present a highly efficient and systematic method of selecting the "optimal split" for decision tree algorithms, called \textit{Superfast Selection}. 
Superfast Selection employs the \textit{prefix-sum} technique to speed up the split selection on a single feature by lowering the time complexity to O(M) (M is the number of examples). 
Additionally, we presented our experiments with Ultrafast Decision Tree (UDT) that uses our Superfast Split method and which can complete training with a time complexity of O(KM$^2$) (K is the number of features), and reported the results on 18 datasets. 
Furthermore, Superfast Selection empowers Ultrafast Decision Tree to handle hybrid features in a consolidated way eliminating the need of pre-encoding for feature-value heterogeneity. 
The result shows that even a lightweight laptop can finish training with UDT on datasets with millions of examples in seconds. 
Finally, \textit{Training Only Once Tuning} helps UDT avoid repetitive training in the tuning process.

The decision tree algorithm with Superfast Selection, Ultrafast Decision Tree, possesses several advantages: 
1) speeds up current applications of decision tree algorithms, and enables them to handle much larger datasets. 
This results in reduced operation costs in time, money, energy, and the cost of purchasing devices for more computational resources; 
2) allows ``resource-constrained devices'' like wearable devices, IoT devices, and mobile phones to have the ability to train machine learning models much more efficiently. 
IoT devices can learn from data and make immediate decisions without needing to send data to servers or cloud for processing \cite{zhang2022federated}. 
This would reduce latency in decision-making processes (crucial for applications requiring instant response such as autonomous vehicles and real-time monitoring systems) and increase autonomy in devices with the ability to train models instantly \cite{merenda2020}. 
IoT devices could particularly benefit in remote or extreme environments without network or human intervention, like deep-sea sensors or rovers; 
3) the instant learning ability allows for decision trees to be used in environments that demand real-time performance, like financial markets \cite{Wang2018}. 
It can be used to instantly adjust pricing models based on real-time market demand, competitor pricing, and inventory levels. 
Instant training would allow for immediate reaction to market changes, maximizing profitability and market competitiveness.


\newpage
\bibliographystyle{plainnat}
\bibliography{mycitations}

\begin{thebibliography}{38}
\providecommand{\natexlab}[1]{#1}
\providecommand{\url}[1]{\texttt{#1}}
\expandafter\ifx\csname urlstyle\endcsname\relax
  \providecommand{\doi}[1]{doi: #1}\else
  \providecommand{\doi}{doi: \begingroup \urlstyle{rm}\Url}\fi

\bibitem[mis()]{misc_statlog_(shuttle)_148}
{Statlog (Shuttle)}.
\newblock UCI Machine Learning Repository.
\newblock {DOI}: https://doi.org/10.24432/C5WS31.

\bibitem[Alpaydin and Kaynak(1998)]{misc_optical_recognition_of_handwritten_digits_80}
E.~Alpaydin and C.~Kaynak.
\newblock {Optical Recognition of Handwritten Digits}.
\newblock UCI Machine Learning Repository, 1998.
\newblock {DOI}: https://doi.org/10.24432/C50P49.

\bibitem[Arzamasov(2018)]{misc_electrical_grid_stability_simulated_data__471}
Vadim Arzamasov.
\newblock {Electrical Grid Stability Simulated Data }.
\newblock UCI Machine Learning Repository, 2018.
\newblock {DOI}: https://doi.org/10.24432/C5PG66.

\bibitem[Becker and Kohavi(1996)]{misc_adult_2}
Barry Becker and Ronny Kohavi.
\newblock {Adult}.
\newblock UCI Machine Learning Repository, 1996.
\newblock {DOI}: https://doi.org/10.24432/C5XW20.

\bibitem[Bishop(2006)]{bishop2006}
Christopher~M. Bishop.
\newblock \emph{Pattern Recognition and Machine Learning (Information Science and Statistics)}.
\newblock Springer-Verlag, Berlin, Heidelberg, 2006.
\newblock ISBN 0387310738.

\bibitem[Blackard(1998)]{misc_covertype_31}
Jock Blackard.
\newblock {Covertype}.
\newblock UCI Machine Learning Repository, 1998.
\newblock {DOI}: https://doi.org/10.24432/C50K5N.

\bibitem[Blanc et~al.(2021)Blanc, Lange, Qiao, and Tan]{blanc2021decision}
Guy Blanc, Jane Lange, Mingda Qiao, and Li-Yang Tan.
\newblock Decision tree heuristics can fail, even in the smoothed setting, 2021.

\bibitem[Breiman(1984)]{breiman1984}
Leo Breiman.
\newblock Classification and regression trees.
\newblock \emph{Wadsworth International Group}, 1984.
\newblock \doi{10.5555/55768}.

\bibitem[Candanedo(2017)]{misc_appliances_energy_prediction_374}
Luis Candanedo.
\newblock {Appliances Energy Prediction}.
\newblock UCI Machine Learning Repository, 2017.
\newblock {DOI}: https://doi.org/10.24432/C5VC8G.

\bibitem[Chen and Guestrin(2016)]{chen2016}
Tianqi Chen and Carlos Guestrin.
\newblock Xgboost: A scalable tree boosting system.
\newblock In \emph{Proceedings of the 22nd ACM SIGKDD International Conference on Knowledge Discovery and Data Mining}, KDD '16, page 785–794, New York, NY, USA, 2016. Association for Computing Machinery.
\newblock ISBN 9781450342322.
\newblock \doi{10.1145/2939672.2939785}.
\newblock URL \url{https://doi.org/10.1145/2939672.2939785}.

\bibitem[Cortez et~al.(2009)Cortez, A.Cerdeira, Almeida, Matos, and Reis]{misc_wine_quality_186}
Paulo Cortez, A.Cerdeira, F.~Almeida, T.~Matos, and J.~Reis.
\newblock {Wine Quality}.
\newblock UCI Machine Learning Repository, 2009.
\newblock {DOI}: https://doi.org/10.24432/C56S3T.

\bibitem[Domingos and Hulten(2000)]{domingos2000}
Pedro Domingos and Geoff Hulten.
\newblock Mining high-speed data streams.
\newblock In \emph{Proceedings of the Sixth ACM SIGKDD International Conference on Knowledge Discovery and Data Mining}, KDD '00, page 71–80, New York, NY, USA, 2000. Association for Computing Machinery.
\newblock ISBN 1581132336.
\newblock \doi{10.1145/347090.347107}.
\newblock URL \url{https://doi.org/10.1145/347090.347107}.

\bibitem[Fanaee-T(2013)]{misc_bike_sharing_275}
Hadi Fanaee-T.
\newblock {Bike Sharing}.
\newblock UCI Machine Learning Repository, 2013.
\newblock {DOI}: https://doi.org/10.24432/C5W894.

\bibitem[for Disease~Control and (CDC)(2022)]{HeartDiseaseIndicators}
Centers for Disease~Control and Prevention (CDC).
\newblock Heart disease health indicators dataset.
\newblock \url{https://www.kaggle.com/datasets/alexteboul/heart-disease-health-indicators-dataset}, 2022.
\newblock Accessed: 2024-05-20.

\bibitem[Freire et~al.(2010)Freire, Veloso, and Barreto]{misc_wall-following_robot_navigation_data_194}
Ananda Freire, Marcus Veloso, and Guilherme Barreto.
\newblock {Wall-Following Robot Navigation Data}.
\newblock UCI Machine Learning Repository, 2010.
\newblock {DOI}: https://doi.org/10.24432/C57C8W.

\bibitem[Kang(2013)]{kang2013}
Hyun Kang.
\newblock The prevention and handling of the missing data.
\newblock \emph{Korean journal of anesthesiology}, 64:\penalty0 402--6, 05 2013.
\newblock \doi{10.4097/kjae.2013.64.5.402}.

\bibitem[Kass(1980)]{kass1980}
Gordon~V. Kass.
\newblock An exploratory technique for investigating large quantities of categorical data.
\newblock \emph{Journal of The Royal Statistical Society Series C-applied Statistics}, 29:\penalty0 119--127, 1980.
\newblock URL \url{https://api.semanticscholar.org/CorpusID:61329067}.

\bibitem[Little(2008)]{misc_parkinsons_174}
Max Little.
\newblock {Parkinsons}.
\newblock UCI Machine Learning Repository, 2008.
\newblock {DOI}: https://doi.org/10.24432/C59C74.

\bibitem[Malerba(1995)]{misc_page_blocks_classification_78}
Donato Malerba.
\newblock {Page Blocks Classification}.
\newblock UCI Machine Learning Repository, 1995.
\newblock {DOI}: https://doi.org/10.24432/C5J590.

\bibitem[Merenda et~al.(2020)Merenda, Porcaro, and Iero]{merenda2020}
Massimo Merenda, Carlo Porcaro, and Demetrio Iero.
\newblock Edge machine learning for ai-enabled iot devices: A review.
\newblock \emph{Sensors}, 20\penalty0 (9), 2020.
\newblock ISSN 1424-8220.
\newblock \doi{10.3390/s20092533}.
\newblock URL \url{https://www.mdpi.com/1424-8220/20/9/2533}.

\bibitem[Nataraj(2018)]{ChurnModelling}
Anand Nataraj.
\newblock Churn modelling dataset.
\newblock \url{https://www.kaggle.com/datasets/man0007/churn-modelling}, 2018.
\newblock Accessed: 2024-05-20.

\bibitem[Neshat et~al.(2023)Neshat, Alexander, Sergiienko, and Wagner]{misc_large-scale_wave_energy_farm_882}
Mehdi Neshat, Bradley Alexander, Nataliia Sergiienko, and Markus Wagner.
\newblock {Large-scale Wave Energy Farm}.
\newblock UCI Machine Learning Repository, 2023.
\newblock {DOI}: https://doi.org/10.24432/C5GG7Q.

\bibitem[Pearson(1900)]{pearson1900}
Karl Pearson.
\newblock On the criterion that a given system of deviations from the probable in the case of a correlated system of variables is such that it can be reasonably supposed to have arisen from random sampling.
\newblock \emph{Philosophical Magazine Series 5}, 50\penalty0 (302):\penalty0 157--175, 1900.
\newblock \doi{10.1080/14786440009463897}.

\bibitem[Quinlan(1986)]{quinlan1986}
J.~Ross Quinlan.
\newblock Induction of decision trees.
\newblock \emph{Machine Learning}, 1\penalty0 (1):\penalty0 81--106, 1986.
\newblock \doi{10.1023/A:1022643204877}.

\bibitem[Quinlan(1993)]{quinlan1993}
J.~Ross Quinlan.
\newblock \emph{C4.5: Programs for Machine Learning}.
\newblock Morgan Kaufmann Publishers Inc., San Francisco, CA, USA, 1993.
\newblock ISBN 1558602402.

\bibitem[Quinlan(1987)]{quinlan1987}
J.R. Quinlan.
\newblock Simplifying decision trees.
\newblock \emph{International Journal of Man-Machine Studies}, 27\penalty0 (3):\penalty0 221--234, 1987.
\newblock ISSN 0020-7373.
\newblock \doi{https://doi.org/10.1016/S0020-7373(87)80053-6}.
\newblock URL \url{https://www.sciencedirect.com/science/article/pii/S0020737387800536}.

\bibitem[R(2022)]{credit_card_fraud}
Dhanush~Narayanan R.
\newblock Credit card fraud.
\newblock \url{https://www.kaggle.com/datasets/dhanushnarayananr/credit-card-fraud}, 2022.

\bibitem[Rajkovic(1997)]{misc_nursery_76}
Vladislav Rajkovic.
\newblock {Nursery}.
\newblock UCI Machine Learning Repository, 1997.
\newblock {DOI}: https://doi.org/10.24432/C5P88W.

\bibitem[Rudin(2019)]{rudin2019}
Cynthia Rudin.
\newblock Do simpler models exist and how can we find them?
\newblock In \emph{Proceedings of the 25th ACM SIGKDD International Conference on Knowledge Discovery \& Data Mining}, KDD '19, page 1–2, New York, NY, USA, 2019. Association for Computing Machinery.
\newblock ISBN 9781450362016.
\newblock \doi{10.1145/3292500.3330823}.
\newblock URL \url{https://doi.org/10.1145/3292500.3330823}.

\bibitem[Rudin et~al.(2021)Rudin, Chen, Chen, Huang, Semenova, and Zhong]{rudin2021interpretable}
Cynthia Rudin, Chaofan Chen, Zhi Chen, Haiyang Huang, Lesia Semenova, and Chudi Zhong.
\newblock Interpretable machine learning: Fundamental principles and 10 grand challenges, 2021.

\bibitem[Sakar and Kastro(2018)]{misc_online_shoppers_purchasing_intention_dataset_468}
C.~Sakar and Yomi Kastro.
\newblock {Online Shoppers Purchasing Intention Dataset}.
\newblock UCI Machine Learning Repository, 2018.
\newblock {DOI}: https://doi.org/10.24432/C5F88Q.

\bibitem[Sarker(2021)]{sarker2021deep}
Iqbal~H Sarker.
\newblock Deep learning: a comprehensive overview on techniques, taxonomy, applications and research directions.
\newblock \emph{SN Computer Science}, 2\penalty0 (6):\penalty0 420, 2021.

\bibitem[Slate(1991)]{misc_letter_recognition_59}
David Slate.
\newblock {Letter Recognition}.
\newblock UCI Machine Learning Repository, 1991.
\newblock {DOI}: https://doi.org/10.24432/C5ZP40.

\bibitem[Stolfo et~al.(1999)Stolfo, Fan, Wenke~Lee, and Chan]{misc_kdd_cup_1999_data_130}
Salvatore Stolfo, Wei Fan, Andreas~Prodromidis Wenke~Lee, and Philip Chan.
\newblock {KDD Cup 1999 Data}.
\newblock UCI Machine Learning Repository, 1999.
\newblock {DOI}: https://doi.org/10.24432/C51C7N.

\bibitem[Velloso and Ugulino(2013)]{misc_weight_lifting_exercises_monitored_with_inertial_measurement_units_273}
Eduardo Velloso and Wallace Ugulino.
\newblock {Weight Lifting Exercises monitored with Inertial Measurement Units}.
\newblock UCI Machine Learning Repository, 2013.
\newblock {DOI}: https://doi.org/10.24432/C5WK63.

\bibitem[Wang et~al.(2018)Wang, Sun, Liu, Cao, and Wang]{Wang2018}
Jia Wang, Tong Sun, Benyuan Liu, Yu~Cao, and Degang Wang.
\newblock Financial markets prediction with deep learning.
\newblock In \emph{2018 17th IEEE International Conference on Machine Learning and Applications (ICMLA)}. IEEE, December 2018.
\newblock \doi{10.1109/icmla.2018.00022}.
\newblock URL \url{http://dx.doi.org/10.1109/ICMLA.2018.00022}.

\bibitem[Yeh(2016)]{misc_default_of_credit_card_clients_350}
I-Cheng Yeh.
\newblock {Default of Credit Card Clients}.
\newblock UCI Machine Learning Repository, 2016.
\newblock {DOI}: https://doi.org/10.24432/C55S3H.

\bibitem[Zhang et~al.(2022)Zhang, Gao, He, Zhang, Krishnamachari, and Avestimehr]{zhang2022federated}
Tuo Zhang, Lei Gao, Chaoyang He, Mi~Zhang, Bhaskar Krishnamachari, and Salman Avestimehr.
\newblock Federated learning for internet of things: Applications, challenges, and opportunities, 2022.

\end{thebibliography}

\end{document}